\pdfoutput=1
\documentclass{article}

\usepackage{multicol}
\usepackage{hyperref}

\usepackage[final]{corl_2019} 

\usepackage{algorithm}
\usepackage[noend]{algpseudocode}

\makeatletter
\def\BState{\State\hskip-\ALG@thistlm}
\makeatother

\usepackage{epsfig}
\usepackage{graphicx}
\usepackage{amsmath}
\usepackage{amssymb}
\usepackage{enumitem}
\graphicspath{{figs/}}
\usepackage[font={footnotesize}]{caption}
\usepackage[dvipsnames]{xcolor}
\usepackage{float}
\usepackage{subcaption}
\usepackage[export]{adjustbox}
\usepackage{tabularx}

\newcommand{\fig}[1]{Fig.~\ref{fig:#1}}
\newcommand{\tabl}[1]{Table~\ref{table:#1}}
\newcommand{\sect}[1]{Sec.~\ref{sec:#1}}

\newcommand{\rulesep}{\unskip\ \vrule\ }

\usepackage{etoolbox}
\newbool{isArxiv}
\booltrue{isArxiv}

\ifbool{isArxiv}{
    \newcommand{\asect}[1]{\ref{sec:appendix.#1}} 
    \newcommand{\afig}[1]{Fig.~\ref{fig:appendix.#1}} 
    
    \thispagestyle{plain}
    \pagestyle{plain}
        
}{ 

    \newcommand{\asect}[1]{\ref{sec:appendix.#1} in \cite{LynchLMP19}}
    \newcommand{\afig}[1]{Appendix Fig.~\ref{fig:appendix.#1} in \cite{LynchLMP19}}

}

\definecolor{ashgrey}{rgb}{0.7, 0.75, 0.71}

\newcommand{\lmpns}{Play-LMP} 
\newcommand{\lmp}{\lmpns\ } 
\newcommand{\gcbcns}{Play-GCBC}
\newcommand{\gcbc}{\gcbcns\ }

\newcommand{\venc}{$V_{enc}$\ }
\newcommand{\cgenc}{$CG_{enc}$\ }
\newcommand{\pilmp}{${\pi_{LMP}}$\ }

\usepackage{amsmath}

\usepackage{amsfonts}

\begin{document}

\title{Learning Latent Plans\\from Play}

\author{
Corey Lynch, Mohi Khansari, Ted Xiao, Vikash Kumar,\\
\textbf{Jonathan Tompson, Sergey Levine, Pierre Sermanet}
\\[0.3em]
Google Brain
}


\maketitle



\vspace{-.3in}

\begin{abstract}
Acquiring a diverse repertoire of general-purpose skills remains an open challenge for robotics. In this work, we propose self-supervising control on top of human teleoperated play data as a way to scale up skill learning. Play has two properties that make it attractive compared to conventional task demonstrations. Play is cheap, as it can be collected in large quantities quickly without task segmenting, labeling, or resetting to an initial state. Play is naturally rich, covering $\sim$4x more interaction space than task demonstrations for the same amount of collection time. To learn control from play, we introduce Play-LMP, a self-supervised method that learns to organize play behaviors in a latent space, then reuse them at test time to achieve specific goals.
Combining self-supervised control with a diverse play dataset shifts the focus of skill learning from a narrow and discrete set of tasks to the full continuum of behaviors available in an environment. We find that this combination generalizes well empirically---after self-supervising on unlabeled play, our method substantially outperforms individual expert-trained policies on 18 difficult user-specified visual manipulation tasks in a simulated robotic tabletop environment. We additionally find that play-supervised models, unlike their expert-trained counterparts, are more robust to perturbations and exhibit retrying-till-success behaviors.
Finally, we find that our agent organizes its latent plan space around functional tasks, despite never being trained with task labels. 
Videos, code and data are available at  {\href{https://learning-from-play.github.io}{learning-from-play.github.io}}
\end{abstract}
\vspace{-.2in}


\begin{figure}[htb]
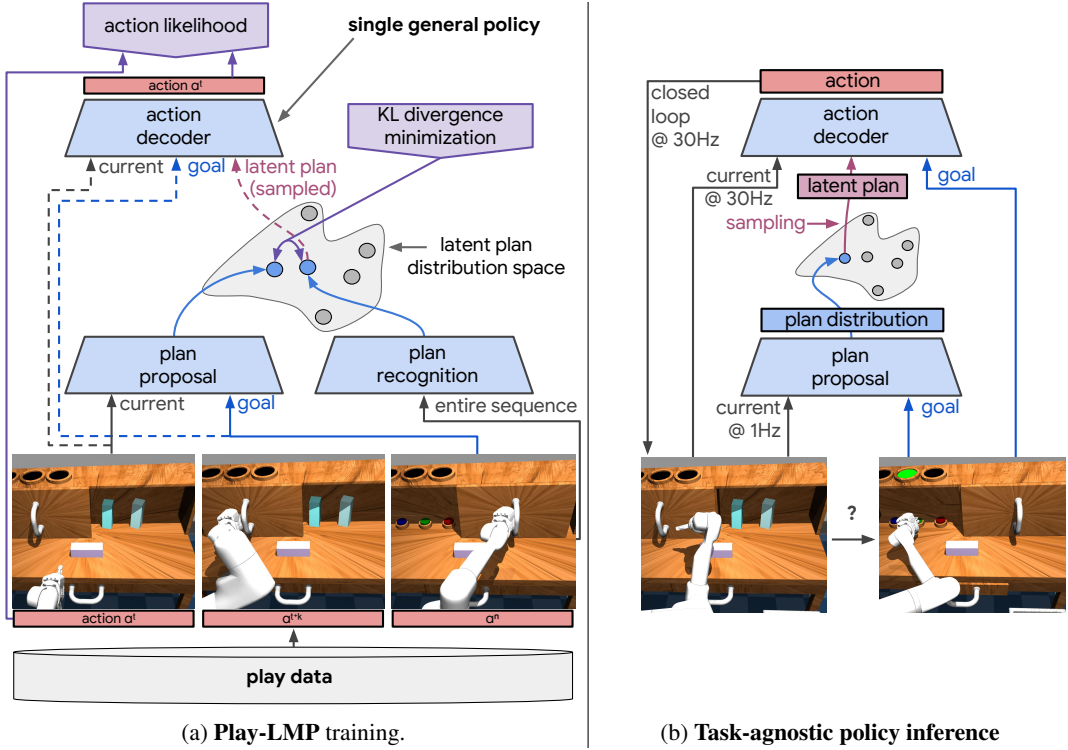

    \begin{subfigure}[t]{.55\textwidth}
        \adjustimage{width=1\linewidth, left}{models/lmp_teaser5}
        \caption{\textbf{\lmpns} training.}
        \label{fig:teaser}
    \end{subfigure}%
    \rulesep    
    \begin{subfigure}[t]{.45\textwidth}
        \adjustimage{width=.9\linewidth, right}{models/lmp_inference5}
        \caption{\textbf{Task-agnostic policy inference}}
        \label{fig:inference}
    \end{subfigure} 
    \caption{\textbf{\lmpns:} A single model that self-supervises control from play data, then generalizes to a wide variety of manipulation tasks. (\subref{fig:teaser}) \textbf{Training}: 1) sample a random window of experience from a memory of play data; 2) train to recognize and organize a repertoire of behaviors executed during play in a \textbf{latent plan space}, 3) train a policy, conditioned on current state, goal state, and a sampled latent plan to reconstruct the actions in the selected window.
    The latent plan space is shaped by two stochastic encoders: plan recognition and plan proposal. Plan recognition takes the entire sequence, recognizing the exact behavior executed. Plan proposal takes the initial and final state, outputting a distribution over all possible behaviors that connect initial state to final. We minimize the KL divergence between the two encoders, making the plan proposal assign high likelihood to behaviors that were actually executed during play.
    (\subref{fig:inference}) \textbf{Inference}: the policy is conditioned on the current state, the goal state (specified by the user) and a latent plan which is sampled once from a plan distribution (inferred from the current and goal states).
    }
\vspace{-.25in}
\end{figure} 

\section{Introduction}
There has been significant recent progress showing that robots can be trained to be competent specialists, learning complex individual skills like grasping (\cite{kalashnikov2018qt}), locomotion, and dexterous manipulation (\cite{haarnoja2018soft}). In this work, we are motivated by the idea of a robot generalist: A single agent capable of learning a wide variety of skills. This remains a challenging open problem in robotics.

Conventionally, obtaining multiple skills involves defining a discrete set of tasks we care about, collecting a large number of labeled and segmented expert demonstrations per task, then training one specialist policy per task in a learning from demonstration
(LfD) \cite{pastor2009learning} scenario. Alternatively, we might turn to reinforcement learning as a means of obtaining a set of skills, which would entail manually designing one reward per task to drive policy learning. Unfortunately, designing reward functions for robotic skills is very challenging, especially when learning from raw observations, typically requiring manually-designed perception systems. Additionally, using reinforcement learning in complex settings such as robotics requires overcoming significant exploration challenges, typically addressed by introducing manual scripting primitives to an unsupervised collection (\cite{ebert2018visual}) that increase the likelihood of behavior with non-zero reward. In general for both paradigms, for each new skill a robot is required to perform, a corresponding, sizeable, and non-transferable human effort must be expended. 

Furthermore, in real world settings, agents will be expected to perform not just a small discrete set of tasks, but rather a wide continuum of behaviors. This presents a challenge for conventional methods---if a slight variation of a skill is needed, e.g. opening a drawer by grasping the handle from the top down rather than bottom up, an entirely new set of demonstrations or reward functions might be required to learn the behavior. To address this, we are motivated by the idea of an agent capable of \emph{task-agnostic control}: the ability to reach \emph{any} reachable goal state from any current state \cite{DBLP:journals/corr/abs-1811-11359}. In this setting, the notion of ``task" is no longer discrete, but continuous---indexed by the pair (current state $s_c$, goal state $s_g$). Learning in this setting can be formalized as the search for a goal-conditioned policy $\pi_{\theta}(a | s_c, s_g)$ (\citet{kaelbling1993learning}).

\begin{figure}[tb]
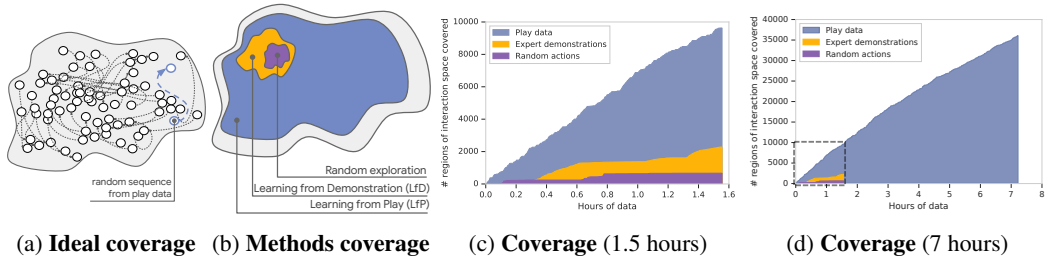

\centering
\scriptsize
    \begin{subfigure}[b]{.19\textwidth}
        \adjustimage{width=1\linewidth, center}{coverage/continuum_density3}
        \caption{\textbf{Ideal coverage}}
        \label{fig:continuum_density}
    \end{subfigure} 
    \begin{subfigure}[b]{.21\textwidth}
        \adjustimage{width=1\linewidth, center}{coverage/continuum3}
        \caption{\textbf{Methods coverage}}
        \label{fig:continuum}
    \end{subfigure} 
    \begin{subfigure}[b]{.29\textwidth}
        \adjustimage{width=1\linewidth, center}{coverage/coverage_limitTrue}
        \caption{\textbf{Coverage} (1.5 hours)}
        \label{fig:coverage_one}
    \end{subfigure}
    \begin{subfigure}[b]{.29\textwidth}
        \adjustimage{width=1\linewidth, center}{coverage/coverage_limitFalse}
        \caption{\textbf{Coverage} (7 hours)}
        \label{fig:coverage_seven}
    \end{subfigure} 
    \caption{\textbf{The continuum of skills and its coverage.} We advocate for learning the full continuum of skills at once rather than discrete ones.
    (\subref{fig:continuum_density}) The ideal coverage is dense and broad over all regions of the space, providing statistical support for all pairs of (current state, goal state).
    (\subref{fig:continuum}) We hypothesize different approaches yield different coverages.
    (\subref{fig:coverage_one}) We observe in real datasets that for the same amount of collection time, play data's coverage largely surpasses that of 18 tasks worth of expert demonstrations and random exploration.
    Unlike the other methods, play data coverage appears to grow linearly with collection time. This prompted us to explore its coverage at larger scales, where we continued to observe the phenomenon. (\subref{fig:coverage_seven}). See details in \asect{coverage}.
    }
\vspace{-.2in}
\end{figure}

To generalize to the widest variety of tasks at test time (indexed by the pair ($s_c$, $s_g$)), it stands that the agent should see the widest variety of ($s_c$, $s_g$) pairs during training, along with actions that connect current and goal states. The ideal dataset to learn task-agnostic control then is both broad and dense in its coverage of the environment's interaction space: \fig{continuum_density}. Unfortunately, it is difficult to obtain datasets with this sort of coverage (\fig{continuum}) in practice. Random exploration, while cheap to collect, is typically insufficiently rich to power the learning of complex manipulation. Expert demonstrations, on the other hand, can be arbitrarily complex but are expensive to collect, and still typically form narrow training distributions over visited states, leading to an empirical ``distribution shift" problem (\citet{ross2011dagger}) at test time.

In this work, we propose an alternative means of obtaining task-agnostic control---self-supervising on top of unlabeled teleoperated \emph{play data}: continuous logs of low-level observations and actions collected while a human teleoperates the robot and engages in behavior that satisfies their own curiosity. We emphasize two properties that make human play data a compelling choice for the basis of learning goal-conditioned control. Play data is \emph{cheap}: Unlike expert demonstrations (\fig{grid_sliding_demo}), play requires no task segmenting, labeling, or resetting to an initial state, meaning it can be collected quickly in large quantities. Play data is \emph{rich}. Play is not random but rather structured by human knowledge of object affordances (e.g. if people see a button in a scene, they tend to press it). This makes play much more discriminate than what can be achieved by random scripting. Unlike task demonstrations, operators are driven by their own curiosity during play, trying multiple ways of achieving the same outcome or exploring new behaviors. In this way, we can expect play to naturally cover an environment's interaction space. In our datasets \fig{coverage_one}, we find empirically that for the same amount of collection time, play indeed covers 4.2 times more regions of the available interaction space than 18 tasks worth of expert demonstration data, and 14.4 times more regions than random exploration.

In \sect{learning_from_play}, we propose two self-supervised methods for learning task-agnostic control from play: Play-GCBC and Play-LMP.

\section{Related Work}

Robotic learning methods generally require some form of supervision to acquire behavioral skills. Conventionally, this supervision either consists of a cost or reward signal, as in reinforcement learning \cite{sutton2018reinforcement,kober2013reinforcement,deisenroth2013survey}, or demonstrations, as in imitation learning \citet{pastor2009learning}. However, both of these sources of supervision require considerable human effort to obtain: reward functions must be engineered by hand, which can be highly non-trivial in environments with natural observations, and demonstrations must be provided manually for each task. When using high-capacity models, hundreds or even thousands of demonstrations may be required for each task (\citet{DBLP:journals/corr/abs-1710-04615,DBLP:journals/corr/RahmatizadehABL17,rajeswaran2017learning,DBLP:journals/corr/DuanASHSSAZ17}).
In this paper, we instead aim to learn general-purpose policies that can flexibly accomplish a wide range of user-specified tasks, using data that is not task-specific and is easy to collect. Our model can in principle use \emph{any} past experience for training, but the particular data collection approach we used is based on human-provided play data.

In order to distill non task-specific experience into a general-purpose policy, we set up our model to be conditioned on the user-specified goal. Goal conditioned policies have been explored extensively in the literature for reinforcement learning \cite{kaelbling1993learning,nair2018visual, andrychowicz2017hindsight,DBLP:journals/corr/abs-1712-00948,DBLP:journals/corr/abs-1711-06006,DBLP:journals/corr/CabiCHDWF17},
as well as for control via inverse models \cite{DBLP:journals/corr/AgrawalNAML16,DBLP:journals/corr/NairCAIAML17,christiano2016transfer,DBLP:journals/corr/abs-1805-01954}.
Learning powerful goal-conditioned policies with reinforcement learning can produce policies with good long-horizon performance, but is difficult in terms of both the number of samples required and the need for extensive on-policy exploration \cite{pinto2015supersizing,levine2017grasping}.
We instead opt to train our model with supervised learning. This introduces a major challenge, since the distribution over actions that can reach a temporally distant goal from the current state based on the data can be highly multimodal. Even single-task imitation models of this sort must contend with multimodality \cite{rahmatizadeh2018vision}, and goal-conditioned models are typically restricted to short and relatively simple tasks, such as pushing \cite{DBLP:journals/corr/AgrawalNAML16}, re-positioning rope \cite{DBLP:journals/corr/NairCAIAML17}, or short-distance navigation \cite{pathakICLR18zeroshot}. We tackle substantially more temporally extended tasks, using our proposed latent plan model, which models the multimodality explicitly using a hierarchical latent variable model. \citet{hausman2018learning} similarly learn a continuous latent space of closely related manipulation skills, instead learning the space with a discrete set of reinforcement learned tasks, defined by per-task rewards.
In contrast to prior work on few-shot learning from demonstration (\cite{finn2017one,wang2017robust}), our method does not require a meta-training phase, any expensive task-specific demonstrations, or a predefined task distribution. In contrast to prior work that uses reinforcement learning (\citet{DBLP:journals/corr/abs-1810-05017}), it does not require any reward function or costly RL phase.
Finally, \citet{DBLP:journals/corr/abs-1810-01257} derive a similar architecture to \lmpns, justifying it in a hierarchical reinforcement learning setting.

\begin{figure*}[tb]
\begin{minipage}[c]{0.6\linewidth}
        \begin{center}
        \adjustimage{width=.9\linewidth, left}{data/playground0}
        \caption{\textbf{The Playground environment.} Details in \asect{playground}}
        \label{fig:playground}
        \end{center}
\end{minipage}
\hfill
\begin{minipage}[c]{0.4\linewidth}
        \begin{center}
        \centerline{\includegraphics[width=.8\linewidth]{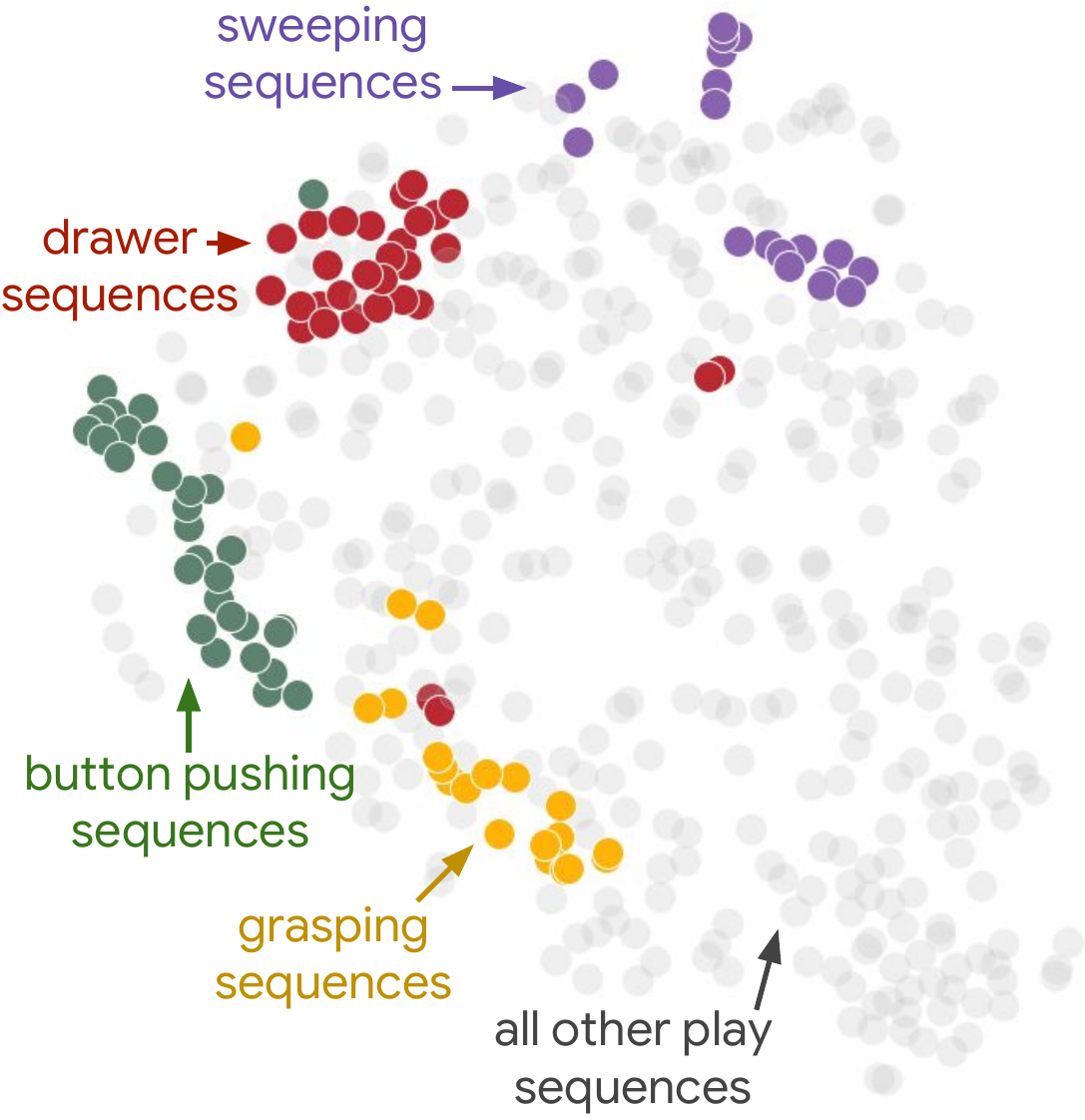}}
        \caption{\textbf{Latent plan space t-SNE} 
        }
        \label{fig:tsne}
        \end{center}
\end{minipage}%
\vspace{-.2in}
\end{figure*}

\begin{figure*}[h]
\begin{center}
\centerline{\includegraphics[width=\linewidth]{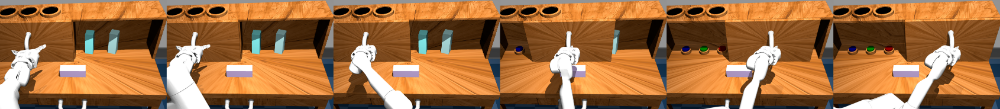}}
\caption{\textbf{Example of a supervised demonstration} sequence labeled and segmented for the "sliding" task.
}
\label{fig:grid_sliding_demo}
\end{center}
\vspace{-.3in}
\end{figure*}

\section{Learning Task-Agnostic Control from Play Data}
\label{sec:learning_from_play}
First we give a definition of the type of data we collect for our experiments and our assumptions about its collection. We create a simulated ``playground environment" (\fig{playground} and \asect{playground}) for play collection and task evaluation. In this environment an 8-DOF simulated robot (arm and gripper) is situated in front of a desk with a sliding door and a drawer. On the desk is a rectangular block and 3 buttons that control lights. See an example of a play sequence in that environment in \asect{play_data}.

\textbf{What is Play?}
We propose play data is generated as follows: A human operator, given the current state of the world $s_c$, formulates a mental image of a goal state they would like to reach next $s_g$, driven by their curiosity or some other intrinsic motivation. For example, in an environment with a ball and cup sitting next to one other, the operator might choose $s_g$ representing ``ball in cup". Given the current state $s_c$ (``ball next to cup"), and goal state $s_g$ (``ball in cup"), the operator considers all the different high-level behaviors $b$ that would achieve the goal. E.g., ``place ball in cup", ``toss ball in cup", ``bounce ball into cup" would all lead to $s_g$. We can consider a prior distribution over all the valid ways of reaching $s_g$ from $s_c$, $p(b | s_c, s_g)$, a behavioral repertoire encoding knowledge of object affordances and environment dynamics. To actually achieve the desired outcome, they sample a single high-level behavior plan from the distribution $b \thicksim p(b | s_c, s_g)$ and execute it, producing the observed stream of low level state and action logs. We emphasize that play is not arbitrary behavior, nor ``random" actions, but rather the very deliberate goal-conditioned behavior that a person engages in under their own direction.

\subsection{Play-Supervised Goal-Conditioned Behavioral Cloning}
We now describe ``play-supervised goal-conditioned behavioral cloning", or Play-GCBC, a method that extracts goal-conditioned policies using self-supervision on top of raw unlabeled play data. Let $D$ be a play dataset, the unsegmented stream of high-dimensional sensory observations and actions logged during teleoperation play. $D$ consists of paired ($O_t$, $a_t$) tuples $D = \{(O_1,a_1),\cdots,(O_T, a_T)\}$. $O_t$ is the set of observations from each of the robot's $N$ sensory channels $\{o^1, ..., o^N\}_t$ at time $t$ and $a_t$ is the logged teleoperation action. In our experiments, $O = \{I, p\}$ consists of $I$, an RGB image observation from a fixed first-person viewpoint, and $p$, the internal 8-DOF proprioceptive state of the agent. See \asect{detailed_architecture} for details.

The key idea behind \gcbc is that a random window of (observation, action) pairs extracted from play describes exactly how the robot got from a particular initial state to a particular final state. Furthermore, it is guaranteed that the final state is reachable from the initial state under the intervening actions. We can exploit this simple structure to create a self-supervised labeling scheme for a goal-conditioned policy, treating the initial state of a random sequence as ``current state", the final state as ``reachable goal state'', and the actions taken as the labels to reproduce.

\textbf{Encoding perceptual inputs:} Since our logs are raw observations, we define one encoder per sensory channel $\Phi = \{E_1, ..., E_N\}$ with parameters $\theta_{\Phi}$, mapping $N$ high-dimensional observations in each $O$ per timestep to one low-dimensional fused state $s_t = concat([E_1(o_1), ..., E_N(o_N)])$. For simplicity we refer to this operation as $s_t \leftarrow \Phi(O_t)$.\\
\textbf{Goal-conditioned policy:} Let $\pi_{GCBC}(a_t| s_t, s_g)$ be a stochastic RNN goal-conditioned policy with parameters $\theta_{GCBC}$, mapping from current state $s_t$ and goal state $s_g$ to the parameters of a distribution over next action $a_t$. We train \gcbc on batches of random play sequences as follows:
For each training batch, and each batch sequence element: we sample a $\kappa$-length sequence of observations and actions $\tau$. We extract the final observation in $\tau$ as the synthetic goal state and encode it. At each timestep $t$ in $\tau$, $\pi_{GCBC}$ takes as input the current state $s_t \leftarrow \Phi(O_t)$ and goal state $s_g$, and maps to the parameters of a distribution over next action $a_t$. Both the encoders and the policy are trained end-to-end to maximize the log likelihood of each action taken during the sampled play sequence.
\begin{equation}
    \mathcal{L}_{GCBC} = -\frac{1}{\kappa} \sum_{t=k}^{k+\kappa} log\big(\pi_{GCBC}(a_t | s_t, s_g)\big)
    \label{eq:gcbc_action_loss}
\end{equation}
We describe the minibatch training pseudo-code in Algorithm~\ref{alg:playgcbc}.






\textbf{Multimodality problem:} A challenge in self-supervising control on top of play is that in general, there are many valid high-level behaviors that might connect the same ($s_c$, $s_g$) pair. This presents multiple counteracting action label trajectories, which can impede learning. Therefore, policies must be expressive enough to model all possible high-level behaviors that lead to the same goal outcome.

\subsection{Play-supervised Latent Motor Plans}
\label{sec:lmp_description}
\textbf{Motivation}.
We propose that unsupervised representation learning is well poised to address the multimodality problem. Consider the following as motivation: if we could learn compact representations of all the different high-level plans that take an agent from a current state to goal state (essentially learning $p(b | s_c, s_g)$) and condition a policy on a single sampled plan, we could convert a multimodal policy learning problem into a unimodal one. That is, a policy previously tasked with a difficult multimodal \emph{plan inference} problem would now be relieved of that problem, and free to use the entirety of its capacity for unimodal \emph{plan execution}. Ideally, individual points in the space correspond to reusable common behaviors executed during teleoperation play. But how do we learn good representations of high-level behavior unsupervised? Furthermore, how would we connect plan representation learning to our main goal of extracting goal-conditioned policies? 

\textbf{Plan Representation Learning Leads to Goal-Conditioned Control}. We turn to the widely influential variational autoencoder (VAE) (\cite{kingma2013auto} framework to learn plans from play. VAEs combine latent representation learning with deep generative models of observed data. Interestingly, we find that by starting with a pure plan representation learning problem and respecting the fact that plans depend on observed current and goal state, the generative decoder part of the model becomes equivalent to a goal and plan-conditioned policy. See \asect{unsup_motivation} for discussion.

We call this method ``Play-supervised Latent Motor Plans", or \lmpns, a unified objective for learning reusable plan representations and task-agnostic control from unlabeled play data. Formally, \lmp is a conditional sequence-to-sequence VAE (seq2seq CVAE) (\citet{NIPS2015_5775}, \citet{45404}), autoencoding random experiences extracted from play through a latent plan space.

As a CVAE, \lmp consists of three components trained end-to-end:
1) Plan recognition: a stochastic sequence encoder, taking a randomly sampled play sequence $\tau$ as input and mapping it to a distribution $q_{\phi}(z|\tau)$ in latent plan space, the learned variational posterior.
2) Plan proposal: a stochastic encoder taking the initial state $s_c$ and final state $s_g$ from the same sequence $\tau$, outputting $p_{\theta}(z|s_c, s_g)$, the learned conditional prior. The goal of this encoder is to represent the full distribution over \emph{all} high-level behaviors that might connect current and goal state, potentially capturing multiple solutions.
3) Plan and goal-conditioned policy: A policy conditioned on the current state $s_c$, goal state $s_g$ and latent plan $z$ sampled from the posterior $q_{\phi}(z|\tau)$, trained to reconstruct the goal-directed actions taken during play, following inferred plan $z$.

Like \gcbcns, \lmp takes as input batches of randomly sampled play sequences $\tau$ and is trained as follows:

\textbf{Plan encoding.}
For each training batch, and each batch sequence element $\tau$:
We first map the sequence of raw observations in $\tau$ to a sequence of encoded states, using perceptual encoders $\Phi$: $\tau* = \Phi(\tau)$.
\venc (``video encoder"), a bidirectional sequence encoder with parameters $\theta_V$, implements the posterior, taking preprocessed $\tau*$ as input and mapping it to the parameters of a distribution in latent plan space: 
$\mu_\phi, \sigma_\phi = {V_{enc}}(\tau*)$.
As is typical with training VAEs, we assume the encoder has a diagonal covariance matrix, i.e. $z \thicksim \mathcal{N}(\mu_\phi, \textrm{diag}(\sigma_\phi^2))$.
Individual latent plans $z$ are sampled from this distribution at training time via the ``reparameterization trick" (\citet{kingma2013auto}) and handed to a latent plan and goal conditioned action decoder (described next) to be decoded into reconstructed actions\footnote{We note that \venc is only used at training time to help learn a latent plan space, and is discarded at test time. While we could in principle use \venc at test time to perform full sequence imitation, in this work we restrict our attention to tasks specified by individual user-provided goal states.}.

\textbf{Plan prior matching.}
We simultaneously extract synthetic ``current" and ``goal" states from the same sequence $\tau$ that \venc just encoded: $s_c \leftarrow \Phi(O_t)$ and $s_g \leftarrow \Phi(O_{t+\kappa})$. We define \cgenc (``current, goal encoder"), to be a feedforward network with parameters $\theta_{CG}$ implementing the learned conditional prior. \cgenc takes concatenated $s_c$ and $s_g$, and outputs the parameters of a distribution in latent plan space: $\mu_\psi, \sigma_\psi = {CG_{enc}}(s_c, s_g)$.
\venc and \cgenc are trained jointly by minimizing the KL divergence between their predicted distributions:
\begin{equation}
\mathcal{L}_{\textrm{KL}} = \textrm{KL}\Big(\mathcal{N}(z|\mu_\phi, \textrm{diag}(\sigma_\phi^2)) ~||~ \mathcal{N}(z|\mu_\psi, \textrm{diag}(\sigma_\psi^2)) \Big)
\label{eq:kl_loss}
\end{equation}
Intuitively, $\mathcal{L}_{\textrm{KL}}$ forces the distribution over plans output by \cgenc to place high probability on actual latent plans recognized during play by \venc.

\textbf{Plan decoding.}
Finally we define \pilmp, a stochastic RNN with parameters $\theta_{LMP}$. \pilmp takes as input current state $s_t$, goal state $s_g$, and a sampled latent plan $z$, and outputs the parameters of a distribution in the agent's action space $A$. The purpose of \pilmp is both to act as a decoder in a representation learning context and a goal and plan-conditioned policy in a task-agnostic control context. We note that by taking plan $z$ as input, the policy is relieved from having to represent multiple high-level plans implicitly, aligning well with the original motivation. We compute the action reconstruction cost as follows: For each timestep $t$ in the input sequence $\tau$, we feed in $s_t \leftarrow \Phi(O_t)$, $s_g$, and $z$ to \pilmp, which outputs the parameters for a probability distribution over observed action $a_t$. We compute the maximum likelihood action reconstruction loss\footnote{We can optionally also have the decoder output state predictions, and add another loss term penalizing a state reconstruction loss.} for each timestep:
\begin{equation}
    \mathcal{L}_\pi = -\frac{1}{\kappa} \sum_{t=k}^{k+\kappa} log\big(\pi_{LMP}(a_t | s_t, s_g, z)\big)
    \label{eq:action_loss}
\end{equation}
Gradients from this loss are backpropagated through \pilmp, the reparameterized sampling operation, \venc, and encoders $\Phi$, optimizing the entire architecture end-to-end.
The full \lmp training objective is:
\begin{equation}
\mathcal{L}_{LMP} = \mathcal{L}_\pi + \beta \mathcal{L}_{\textrm{KL}}
\label{eq:total_loss}
\end{equation}
Note that following \citet{higgins2016beta}, we introduce a weight $\beta$, controlling $\mathcal{L}_{\textrm{KL}}$'s contribution to the total loss. Setting $\beta$ \textless 1 was sufficient to avoid ``posterior collapse'' (\citet{45404}), a commonly identified problem in VAE training in which an over-regularized model combined with a powerful decoder tends to ignore the latent variable $z$. 
We describe the full \lmp minibatch training pseudocode in Algorithm~\ref{alg:lmp}.

\textbf{Task-agnostic control at test time}.
Here we describe how \lmp solves user-provided manipulation tasks at test time. At the beginning of each test episode, the agent starts in some current state $O_c$ and receives a perceptual human-provided goal $O_g$. Both are encoded in state space ($s_c \leftarrow \Phi(O_c)$, $s_g \leftarrow \Phi(O_g)$), concatenated, and fed to the learned conditional prior, \cgenc, which outputs a distribution over high-level latent behavior plans $z$ that should take the agent from $s_c$ to $s_g$. The agent samples a single plan $z$ from the distribution, then decodes it in closed loop in the environment. At each timestep of the decoding, the agent feeds ($s_t$, $s_g$, $z$) to \pilmp, a low level action is sampled $a_t \thicksim \pi_{LMP}(a_t | s_t, s_g, z)$. We allow the agent to ``replan" by inferring and sampling new latent plans every $\kappa$ timesteps (matching the average planning horizon it was trained with). In our experiments, our agent gets observations and takes low-level actions at $30$hz. We set $\kappa$ to $32$, meaning that the agent replans at roughly $1$hz. See \fig{inference} for details.


\vspace{-.1in}
\section{Experiments}
\label{sec:experiments}
\vspace{-.1in}

In our experiments, we aim to answer the following questions: 1) Can a single play-supervised policy generalize to a wide variety of user-specified visual manipulation tasks, despite not being trained on task-specific data? 2) Are play-supervised models trained on cheap to collect play data (LfP) competitive with specialist models trained on expensive expert demonstrations for each task (LfD)? 3) Does decoupling latent plan inference and plan decoding into independent problems, as is done in \lmpns, improve performance over goal-conditioned Behavioral Cloning (Play-GCBC), (which does no explicit latent plan inference)?

\textbf{Tasks and Dataset:} We define 18 visual manipulation tasks (see \asect{tasks}) in the same environment that play was collected in (\fig{playground} and \asect{playground}). To compare our play-supervised models to a conventional scenario, we collect a training set of 100 expert demonstrations per task in the environment, and train one behavioral cloning policy (\textbf{BC}, details in \asect{bc}) on the corresponding expert dataset. This results in 1800 demonstrations total or $\sim$1.5 hours of expert data. We additionally train a single multi-task behavioral cloning baseline conditioned on state and task id, \textbf{Multitask BC} (\citet{rahmatizadeh2018vision}), trained on all 18 BC expert demonstration datasets.
We collect play datasets (example in \asect{play_data}) of various sizes as training data for \textbf{\lmpns} and \textbf{\gcbcns}, up to $\sim$7 hours total. We define two sets of experiments over these datasets: \textbf{pixel experiments}, where we study the multi-task visual manipulation problem, and \textbf{state experiments}, where we ignore the visual representation learning problem and provide all models with ground truth states (positions and orientations of all objects in the scene) as observations. The motivation of the state experiments is to understand the how all methods compare on the control problem independent of visual representation learning, which could potentially be improved independently via other self-supervised methods e.g. \citet{Sermanet2017TCN}.

\begin{figure*}[h]
    \vspace{-.2in}
    \begin{subtable}[b]{.5\textwidth}
        \begin{center}
        \setlength{\tabcolsep}{0.3em}
        \footnotesize
            \begin{tabular}{l|c|c|c}
                                & \multicolumn{2}{c|}{\textbf{training data}}                      &        \\
                \textbf{Method} & \textbf{labels} & \textbf{input} & \textbf{success \%}                \\
                \hline
                \hline
                BC & labeled & pixels & $66.5\% \pm 12.1$ \\                
                \gcbc (ours) &  unlabeled & pixels & $58.7\% \pm 11.6$ \\
                \lmp (ours) &   unlabeled & pixels & $\textbf{69.4\%} \pm 10.8$ \\
\hline                
                BC & labeled & states & $70.3\%$ \\
                Multitask BC  & labeled & states & $66.2\%$ \\
                \gcbc (ours) & unlabeled & states & $77.9\%$ \\
                \lmp (ours)  & unlabeled & states & $\textbf{85.5\%}$ \\
            \end{tabular}
        \vspace{.3in}
        \caption{\textbf{18-task success.}}
        \label{table:debi}
        \end{center}
    \end{subtable}%
    \begin{subfigure}[b]{.5\textwidth}
        \begin{center}
        \adjustimage{width=.95\linewidth, right}{xp/robustness}
        \caption{\textbf{Robustness to variations.}}
        \label{fig:robustness}
        \end{center}
    \end{subfigure}
    \vspace{-.2in}
    \caption{\textbf{Quantitative task success and robustness}. (\subref{table:debi}) Play-LMP consistently outperforms the baselines, whether trained on groundtruth states or directly on pixels. Success is reported with confidence intervals over 3 seeded training runs for pixel experiments. (\subref{fig:robustness}) models trained on play data are more robust to perturbations to the initial position. See \sect{experiments} for details.}
\end{figure*}

\begin{figure*}[h]
    \vspace{-.2in}
    \begin{subfigure}[t]{.5\textwidth}
        \begin{center}
        \centerline{\includegraphics[width=\linewidth]{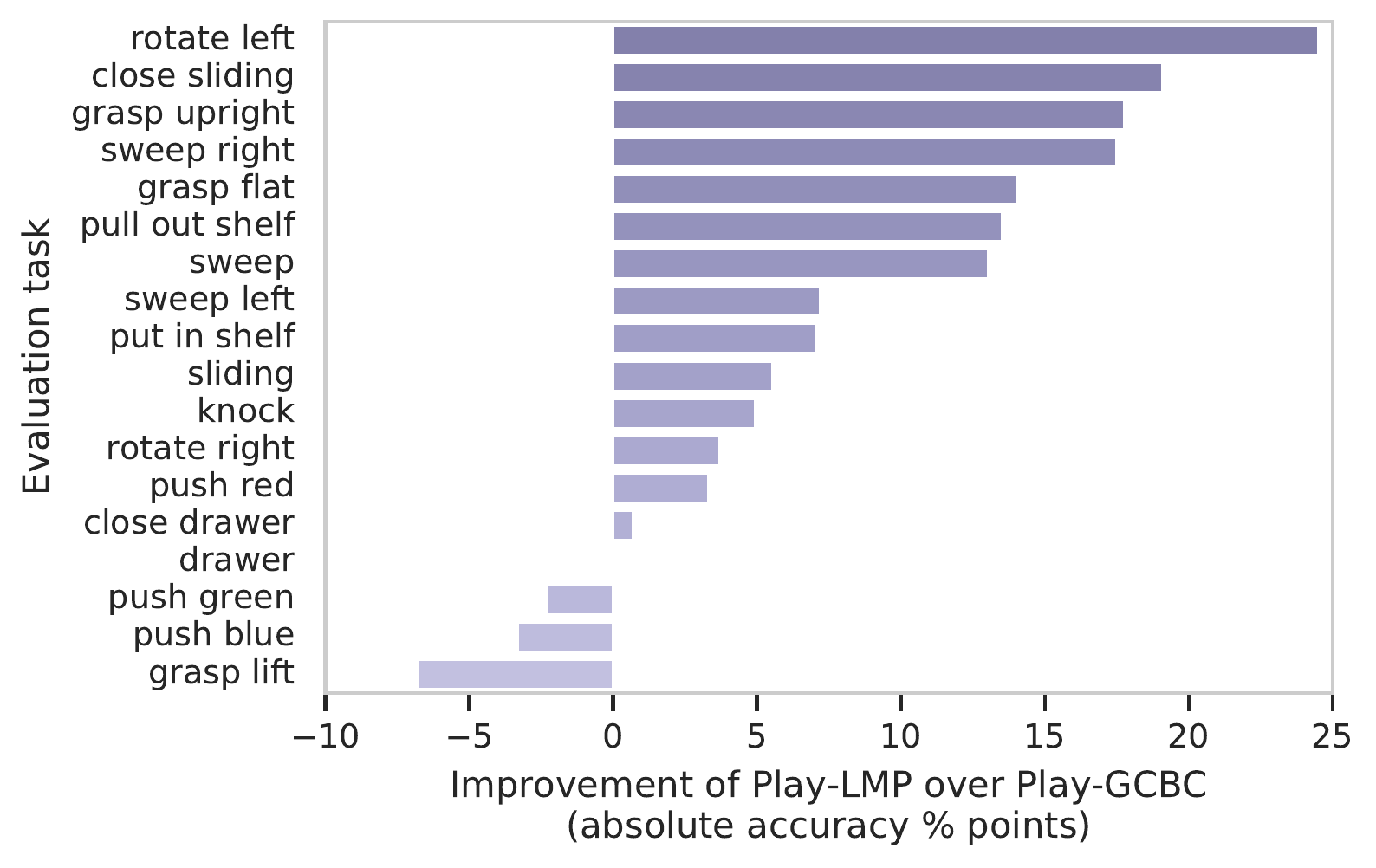}}
        \label{fig:differences_Play-LMP_over_Play-GCBC}
        \end{center}
    \end{subfigure}%
    \begin{subfigure}[t]{.5\textwidth}
        \begin{center}
        \centerline{\includegraphics[width=\linewidth]{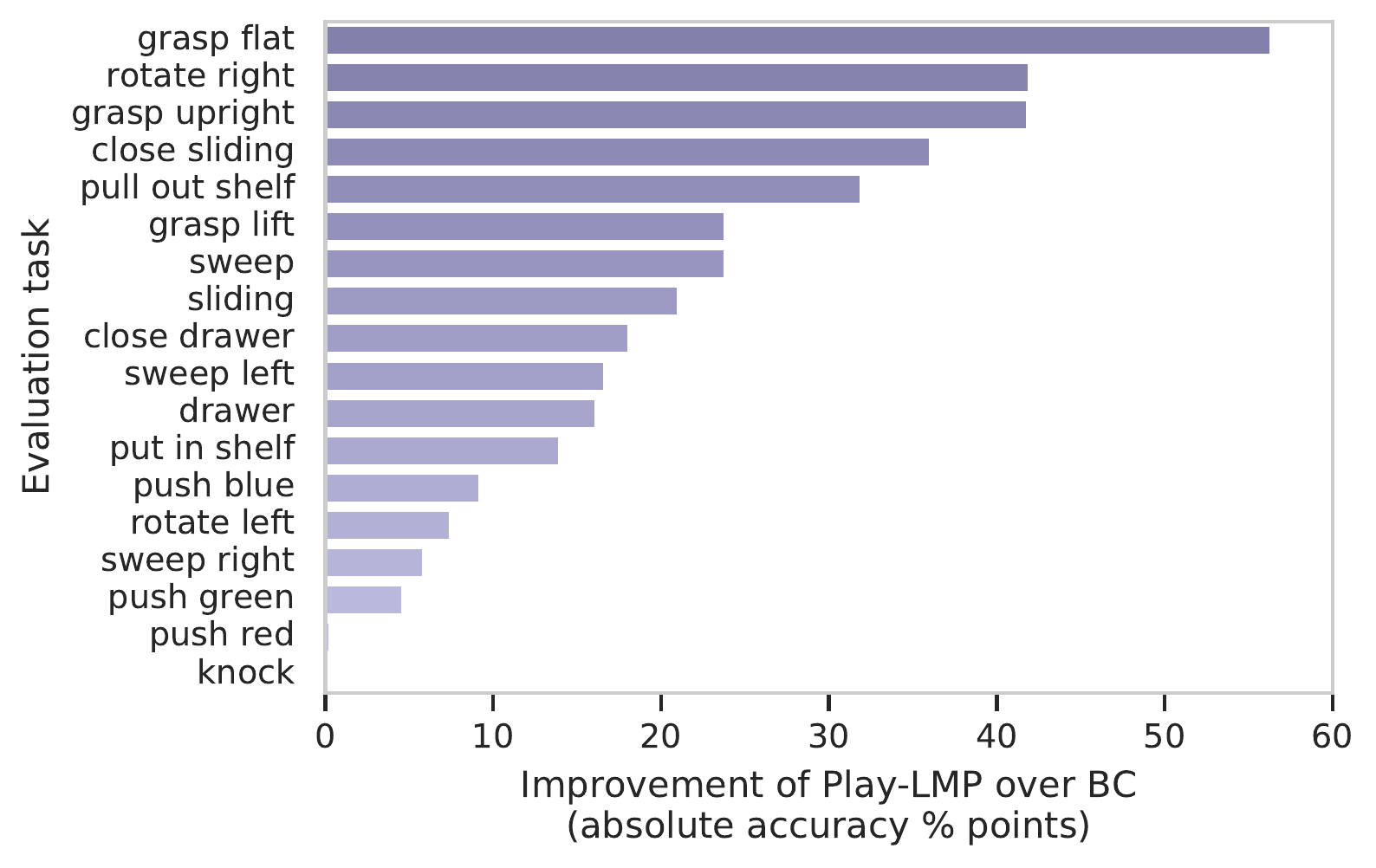}}
        \label{fig:differences_Play-LMP_over_BC}
        \end{center}
    \end{subfigure}
    \vspace{-.2in}
    \caption{\textbf{Improvement per task of Play-LMP} over Play-GCBC (left) and BC baselines (right), in absolute percentage points of accuracy (model trained on states).}
    \label{fig:differences}
\end{figure*}

\textbf{Task success with play-supervision:} We present our main findings for both experiments in \tabl{debi}. First, we find that despite not being trained on task labels, a single \lmp policy outperforms the 18 specialized and supervised BC models (answering experimental questions 1 and 2).
Additionally, we find that the decoupling happening in \lmp compared to \gcbc is beneficial and yields systematic improvements in performance.
We report in \fig{differences} the absolute improvement per task in percentage points of \lmp over the baselines, with up to 50 points of improvement.

\begin{figure*}[h]
\vspace{-.15in}
\begin{center}
\centerline{\includegraphics[width=\linewidth]{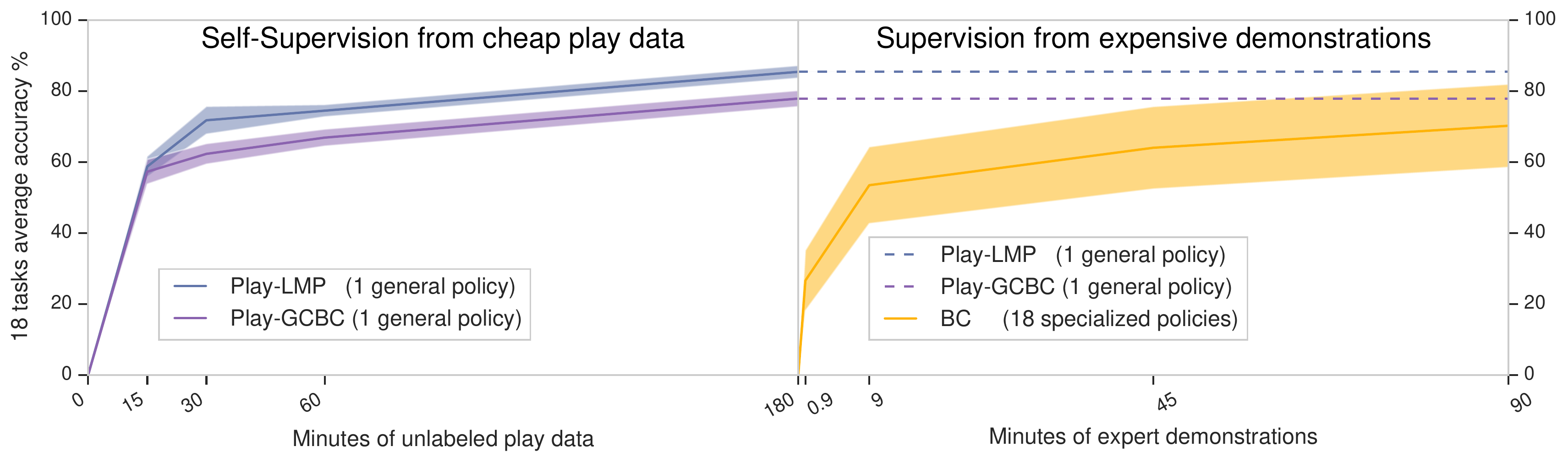}}
\caption{\textbf{18-tasks average success} for self-supervised models trained on various amounts of cheap play data (left) vs. expert-trained models trained on expensive task demonstrations (right).
A single task-agnostic \lmp policy, trained on unlabeled play data generalizes with $85.5\%$ success to the 18 test-time tasks with no finetuning, outperforming a collection of 18 expert-trained BC policies who reach $70.3\%$ average success. This holds true even when Play-LMP is artificially restricted to only 30 minutes of play data ($71.8\%$), despite play being easier and cheaper to collect than expert demonstrations.
These data ablation numbers were obtained from models trained on ground truth state observations. Shaded regions indicate 95\% confidence intervals over 20 rollouts. See \sect{experiments} for details.
}
\label{fig:debi}
\end{center}
\vspace{-.3in}
\end{figure*}

\textbf{Scalability:} 
We see in \fig{debi} that even when trained on only 30 minutes of play data, individual Play-LMP policies outperform 18 BC policies trained on 90 minutes of expert task-specific demonstrations. We feel this highlights the scalability and generality of the approach---that models trained only on random windows extracted from play are prepared for specific tasks presented to them at test time. We believe this comparison is fair for two reasons: 1) the baseline gets 3x more training data, 2) the baseline training data consists of curated task-specific demonstrations of optimal behavior, whereas there is no guarantee that 30 minutes of play data contains optimal task demonstrations.

\textbf{Robustness:} In \fig{robustness}, we find that models trained on play data (Play-LMP and Play-GCBC) are significantly more robust to perturbations than the model trained on expert demonstrations only (BC), a phenomenon we attribute to the inherent coverage properties of play data over demonstration data. More details in \asect{robustness}.

\textbf{Unsupervised task discovery:} We investigate the latent plan spaced learned by \lmpns, seeing whether or not it is capable of encoding task information despite not being trained with task labels. In \fig{tsne} we embed 512 randomly selected windows from the play dataset as well as all validation task demonstrations, using the $\Phi$ plan recognition model. Surprisingly, we find that despite not being trained explicitly with task labels, \lmp appears to organize its latent plan space functionally. E.g. we find certain regions of space all correspond to drawer manipulation, while other regions correspond to button manipulation.

\textbf{Emergent Retrying:} We find qualitative evidence that play-supervised models, unlike models trained solely on expert demonstrations, make multiple attempts to retry the task after initial failure. See \asect{retry}.

\vspace{-.2in}
\section{Conclusion}
\label{sec:conclusion}
\vspace{-.1in}
In this work, we advocate for learning the full continuum of tasks using unlabeled play data, rather than discrete tasks using expert demonstrations.
We introduce a self-supervised plan representation learning algorithm able to discover task semantics despite never seeing any task labels. By learning to generate actions for its task-agnostic policy, the model is able to train an entire deep sensory stack from scratch. We showed that play brings scalability to data collection, as well as robustness to the models trained with it. We explore the setting where play data and test-time tasks are defined over the same playroom environment. Future work includes exploring whether generalization is possible to novel objects or novel environments, as well as exploring the effects of imbalance in play data distributions as discussed in \asect{limitations}.

\textbf{Acknowledgements}\\
\footnotesize{
We thank Michael Wu for designing and iterating on our playground environment. We would also like to thank Karol Hausmann, Ben Poole, Eric Jang for many fruitful discussions on the design and training of our model.
}

\bibliographystyle{plainnat}
\renewcommand{\bibfont}{\footnotesize}
\bibliography{references}

\newpage
\clearpage

\appendix
\section{Appendix}

\subsection{Theoretical Motivation}
\subsubsection{Unsupervised Representation Learning of Plans and Control from Play}
\label{sec:appendix.unsup_motivation}
We describe an unsupervised representation learning setup and show that it can be repurposed for goal-conditioned control.
In the typical unsupervised representation learning setup, we let $p_{data}(x)$ be the true underlying process generating $x$ $\in$ $X$, and $D$ be a dataset of i.i.d. samples from $p_{data}(x)$. A common approach for inducing feature learning is to consider the joint distribution $p(x,z)$ over $(x,z)$, where $x$ $\in$ $X$ are points in the observed data space and $z$ $\in$ $Z$ are points in a latent space. $z$ are typically assumed to have generated $x$.
In our setting, we can consider $x$ to be entire state-action trajectories $\tau$ of average length $\kappa$, sampled from a play dataset. Since our $x$ are entire behavior trajectories, and $z$ is assumed to have generated $x$, we call our $z$ ``latent motor plans", the idea being that ``unobserved plans generate observed goal-directed behavior".
How do we actually go about learning good plans unsupervised from unlabeled data? We can consider a maximum likelihood based approach, where we parameterize the joint $p_{\theta}(x,z)$ and maximize the marginal log likelihood of the observations: $\log p_{\theta}(x)$. 
When $z$ is a continuous variable, marginalization becomes intractable. We can sidestep this issue by turning to stochastic gradient variational Bayes (SGVB) (\citet{kingma2013auto}) framework, which optimizes a surrogate objective function: the variational lower bound of the marginal log-likelihood.
\begin{equation}
\log p_{\theta}(x) \geq -\textrm{KL}\big(q_{\phi}(z|x) ~||~ p_{\theta}(z)\big) + \mathbb{E}_{q_{\phi}(z|x)}\left[ \log p_{\theta}(x|z) \right]
\end{equation}
SGVB replaces an intractable true posterior $p_{\theta}(z|x)$ with an learned approximate inference distribution $q_{\phi}(z|x)$, also known as a ``recognition" model---given a data point $x$, it produces a distribution over the possible values of the code $z$ from which the data point $x$ could have been generated.
Returning to our setting, we ultimately interested in extracting \emph{goal-conditioned} control from play, like in \gcbcns. As a reminder of our assumed generative process: we assume the observed tele-operation logs between the first state $s_c$ and the synthetic goal state $s_g$ in a randomly sampled window $\tau$ are generated as the result of the execution of a high level behavior $b$, sampled from a human operators (unobserved) behavioral repertoire---$p(b|s_c,s_g)$. Since we intend to learn the latent $z$ that matches the unobserved $b$, it makes sense that we similarly condition our learned $z$ on the same observed information: $(s_c, s_g)$.
We can show that when we do this, instead considering the conditional joint probability distribution $p(x,z|c)$ when undergoing representation learning, we can recover \emph{both} representation learning and goal-conditioned control. That is: VAE learns both representations and a generative model of data. By respecting the fact that our representations correspond to plans, and plans depend on current and goal state, the generative part becomes goal-conditioned control.
First, we replace the standard maximum likelihood representation learning objective with the maximum conditional log likelihood: $\log p_{\theta}(x|c)$, where $c$ is the current state and synthetic goal state context we extract from each window sampled from play, $c=(s_c, s_g)$. We note that by substituting actions in $\tau$ for $x$ and $(s_c, s_g)$ for $c$, our new representation learning objective becomes $\log p_{\theta}(a|s_c, s_g)$, equivalent to our control learning objective in \ref{eq:gcbc_action_loss}. Our new tractable representation learning objective is to maximize the variational lower bound on the conditional log likelihood, also referred to as the conditional variational autoencoder or CVAE (\citet{NIPS2015_5775}):
\begin{equation}
\log p_{\theta}(x|c) \geq -\textrm{KL}\big(q_{\phi}(z|x,c) ~||~ p_{\theta}(z|c)\big) + \mathbb{E}_{q_{\phi}(z|x,c)}\left[ \log p_{\theta}(x|z,c) \right]
\end{equation}
We note that this model implies a formal conditional generative process of our data that matches our earlier conceptual one:
For each observed window of state action pairs $x$ of size $\kappa$ sampled from play dataset $D$: 1) Given an observed context $c \leftarrow (s_c$, $s_g$), the current state $s_c$ and goal state $s_g$, 2) Draw latent plan $z$ from conditional prior distribution $z \thicksim p_{\theta}(z|c)$. Note this matches our concept of ``operator drawing high-level plan of how to reach goal from a behavioral repertoire" $b \thicksim p(b|s_c, s_g)$. 3) Draw $x \thicksim p_{\theta}(x|c,z)$, the sequence of intervening states and actions between $s_c$ and $s_g$ according to context and plan-conditioned distribution. Note that this is equivalent to a goal and plan-conditioned policy $\pi_{\theta}(a_t|s_c, s_g, z)$.

We see that from the objective that this leaves us with three modules to implement: the recognition network $q_{\phi}(z|x,c)$, the (conditional) prior network $p_{\theta}(z|c)$, and the generation network $p_{\theta}(x|z,c)$. We now substitute back in the data variables obtained by self-supervised mining of windows from play to define each of \lmpns's modules:
\begin{itemize}
\item $q_{\phi}(z|\tau) \leftarrow q_{\phi}(z|x,c)$. The learned variational posterior becomes a ``plan recognition" network, recognizing which region of latent plan space an observed observation-action sequence belongs to. Note we simplify $q_{\phi}(z|x,c)$ to $q_{\phi}(z|\tau)$ by pointing out that $x$, the intervening state-action sequence between $s_c$ and $s_g$ combined with $c=(s_c,s_g)$ results in the full sequence $\tau$.
\item $p_{\theta}(z|s_c, s_g) \leftarrow p_{\theta}(z|c)$. The learned conditional prior becomes a ``plan proposal" network mapping from current and goal state to a distribution over high level latent plans connecting them.
\item $\pi(a_t|s_c, s_g,z) \leftarrow p_{\theta}(x|z,c)$ The plan and goal conditioned generative network becomes a plan and goal-conditioned policy.
\end{itemize}

\begin{algorithm}[tb]
\footnotesize
  \caption{Training \gcbcns}
  \label{alg:playgcbc}
  \begin{algorithmic}[1]
  \State {\bfseries Input:} Play data $D: \{(s_1,a_1),\cdots,(s_T, a_T)\}$
  \State {\bfseries Input:} Window bounds: $\kappa_{low}$, $\kappa_{high}$
  \State Randomly initialize model parameters $\theta = \{\theta_{GCBC}, \theta_{\Phi}\}$.
  \While{not done}:
  \State Sample a sequence length $\kappa \thicksim U(\kappa_{low}, \kappa_{high})$
  \State Sample a sequence $\tau = \{(O_{t:t+\kappa}, O_{t:t+\kappa})\} \thicksim D$
  \State Set encoded goal state: $s_g \leftarrow \Phi(O_{t+\kappa})$
  \State Compute action loss\newline
  $\textrm{~~~~~~~~~} \mathcal{L}_{GCBC} = -\frac{1}{\kappa} \sum_{t=k}^{k+\kappa} log\big(\pi_{GCBC}(a_t | \Phi(O_t), s_g)\big)$
  \State Update $\theta$ by taking the gradient step to minimize\newline $\textrm{~~~~~}\mathcal{L}_{GCBC}$.
  \EndWhile
  \end{algorithmic}
\end{algorithm}

\begin{algorithm}[tb]
\footnotesize
  \caption{Training \lmpns}
  \label{alg:lmp}
  \begin{algorithmic}[1]
  \State {\bfseries Input:} Play data $\mathcal{D}: \{(s_1,a_1),\cdots,(s_T, a_T)\}$
  \State Randomly initialize model parameters $\theta = \{\theta_{V}, \theta_{CG}, \theta_{{\pi}{LMP}} , \theta_{\Phi}\}$
  \While{not done}:
  \State Sample a sequence $\tau = \{(O_{t:t+\kappa}, a_{t:t+\kappa})\} \thicksim \mathcal{D}$
  \State Map raw observations in $\tau$ to encoded states: $\tau* = \Phi(\tau)$
  \State Map encoded sequence to plan space: $\mu_\phi, \sigma_\phi = {V_{enc}}(\tau*)$
  \State Set current and goal state: $s_i \leftarrow \Phi(O_t),~s_g \leftarrow \Phi(O_{t+\kappa})$
  \State Map encoded (current, goal) to plan space: $\mu_\psi, \sigma_\psi = CG_{enc}(s_t, s_g)$
  \State Compute KL loss using Eq.~\ref{eq:kl_loss}.
  \State Compute action loss using Eq.~\ref{eq:action_loss}.
  \State Update $\theta$ by taking a gradient step to minimize Eq.~\ref{eq:total_loss}.
  \EndWhile
  \end{algorithmic}
\end{algorithm}

\subsubsection{Behavior Cloning}
\label{sec:appendix.bc}
We train one behavioral cloning policy $\pi_{\theta}(a|s)$ per task for each of our 18 tasks. All policy architectures---BC, GCBC, and LMP---have the same architecture: an RNN with 2 hidden layers of size 2048 each, mapping inputs to the parameters of MODL distribution on quantized actions.

\subsection{Architecture Details}
\label{sec:appendix.detailed_architecture}

In \afig{detailed_architecture} we show the layers with their sizes and depths of different sub-networks used in the model: the vision network, plan recognition network, plan proposal network and the policy network. All parameters used by the policy are indicated in green. All inputs given by the environment are indicated in purple. The networks activation maps are displayed in blue.

\textbf{Observation space}
We consider two types of experiments: pixel and state experiments. In the pixel experiments, observations consist of ($I$,$p$) pairs of 299x299x3 RGB images and internal proprioceptive state. Proprioceptive state is the 8 dimensional position, orientation, and gripper elements of the end effector, described below.

In the state experiments, observations consist of the 8-d proprioceptive state, the position and euler angle orientation of the block, and a continuous 1-d sensor describing: {door open amount, drawer open amount, red button pushed amount, blue button pushed amount, green button pushed amount}.

\textbf{Action space}
Our 8-DOF agent's action space state consists of: 3 cartesian coordinates for the position of its end effector, 3 Euler angles representing its end effector orientation, and 2 angles representing its gripper. During training we quantize each action element into 256 bins. All stochastic policy outputs are represented as mixtures of discretized logistic distributions over quantization bins \citet{DBLP:journals/corr/SalimansKCK17}.

\textbf{Encoding perceptual inputs}
In the image experiments, we define a convolutional image embedder, described in \afig{detailed_architecture} to encode image sensory streams. Proprioceptive sensory streams are not transformed by an encoding network, but rather just zero mean, unit variance normalized.

Similarly in the state experiments, we take normalized ground truth position and orientation data as inputs to the models, defining no additional encoding.

\textbf{Goals}
In the image experiments, only the output of the visual embedder is treated as goal state, i.e. not the proprioceptive state. This allows more general goal-specification, e.g. the ability to provide the agent with just a goal image to reach and not also need to provide the internal joint state to reach.

\begin{figure*}[h]
\begin{center}
\centerline{\includegraphics[width=\linewidth]{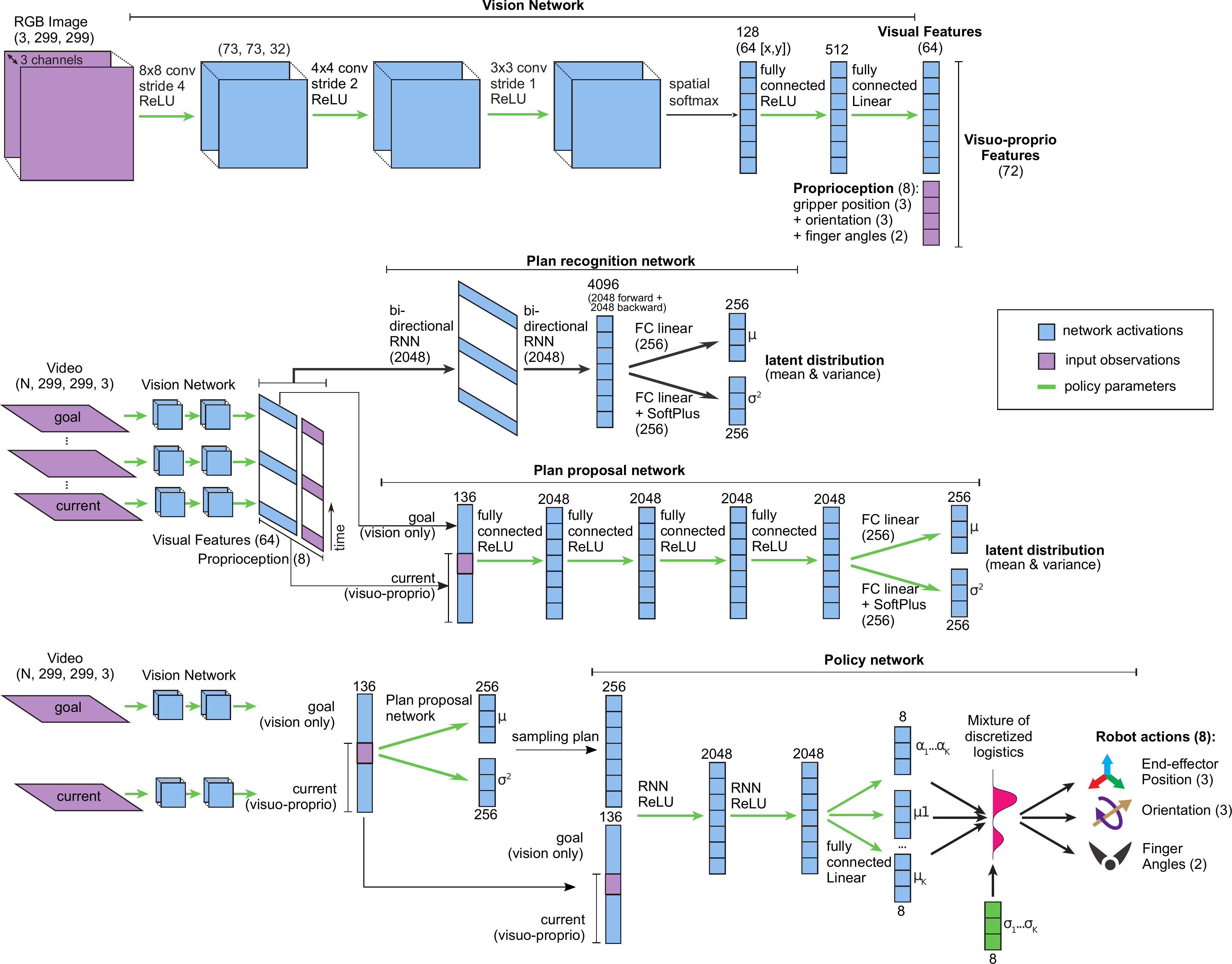}}
\caption{\textbf{Detailed architecture of Play-LMP.}
}
\label{fig:appendix.detailed_architecture}
\end{center}
\end{figure*}

\subsection{Experimental Details}

\begin{figure*}[h]
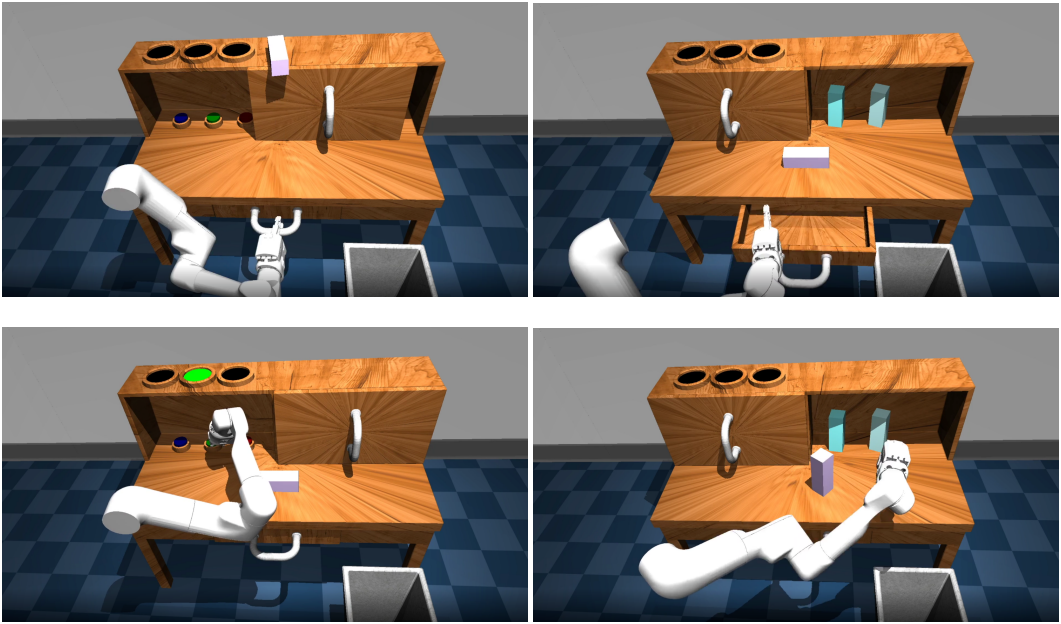

\begin{subfigure}[b]{0.5\linewidth}
        \begin{center}
        \adjustimage{width=1\linewidth, left}{data/playground1}
        \label{fig:playground1}
        \end{center}
\end{subfigure}
\hfill
\begin{subfigure}[b]{0.5\linewidth}
        \begin{center}
        \adjustimage{width=1\linewidth, left}{data/playground2}
        \label{fig:playground2}
        \end{center}
\end{subfigure}
\\
\begin{subfigure}[b]{0.5\linewidth}
        \begin{center}
        \adjustimage{width=1\linewidth, left}{data/playground3}
        \label{fig:playground1}
        \end{center}
\end{subfigure}
\hfill
\begin{subfigure}[b]{0.5\linewidth}
        \begin{center}
        \adjustimage{width=1\linewidth, left}{data/playground4}
        \label{fig:playground2}
        \end{center}
\end{subfigure}%
        \caption{\textbf{The Playground environment.} This environment contains an actionable door, drawer, buttons and objects.}
\end{figure*}

\subsubsection{Playground Environment}
\label{sec:appendix.playground}

We created a simulated ``playground environment" with enough diversity that it can be used for general playing as well as evaluating specific tasks.
An example of it can be seen in \fig{playground}. In this environment an 8-dof simulated robot (arm and gripper) is situated in front of a desk with a sliding door and a drawer.
On the desk is a rectangular block and 3 buttons that control lights. The scene also has rectangular objects that can be interacted with as well as a trash bin on the floor.
In that environment we can quantitatively determine success on specific tasks (defined in \asect{tasks}). We use these success (or reward) functions for evaluation purposes only.

\subsubsection{Play Data}
\label{sec:appendix.play_data}

In \afig{grid_playground} we show an example of a play sequence, by displaying frames sampled every second from a same sequence and
ordered from left to right and top to bottom.
We see the human operator engaging in self-guided interaction with a rectangular object through VR teleoperation.
In this case, the operator chooses to pick up the object, push it around, uses it to push the door to the left, drops the object inside the cabinet, then finally drops the object off the table.
Our play dataset consists of 7 hours of unscripted continuous play similar to this sequence.
Note that subsequences could be considered task demonstrations, e.g. when the agent places the block inside the shelf. Although, they might not necessarily be expert demonstrations, but rather incompletely functional, containing misses, inefficient behavior, etc. Also note that not all the behaviors observed during play are evaluated, e.g. when the agent drops the object off the table or opens the door with the block.

\begin{figure*}[h]
\begin{center}
\centerline{\includegraphics[width=\linewidth]{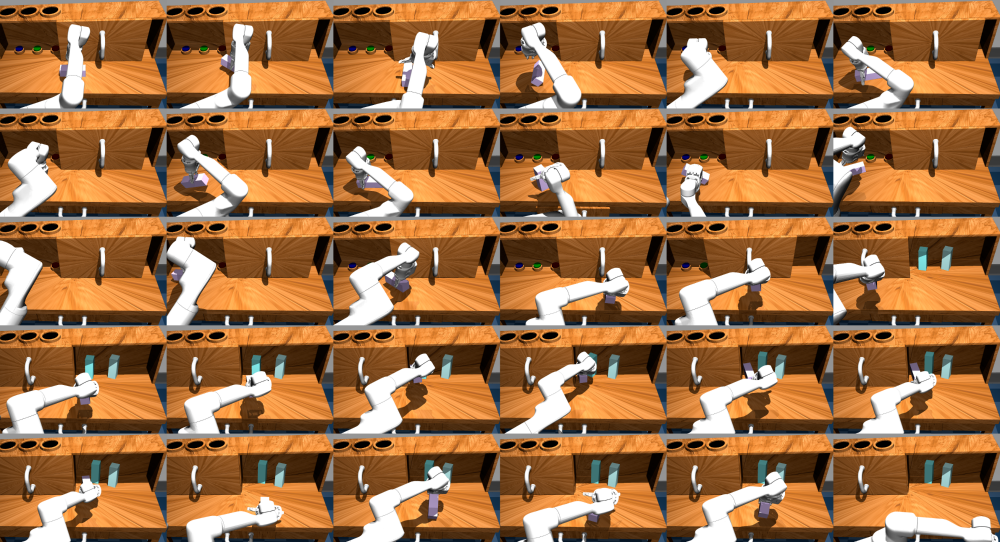}}
\caption{\textbf{Example of ``play" data.}}
\label{fig:appendix.grid_playground}
\end{center}
\end{figure*} 

\subsubsection{Tasks Descriptions}
\label{sec:appendix.tasks}

Here we list the 18 tasks we use to evaluate each method at test time.
\begin{itemize}
\item Grasp lift: Grasp a block out of an open drawer and place it on the desk surface.
\item Grasp upright: Grasp an upright block off of the surface of the desk and lift it to a desired position.
\item Grasp flat: Grasp a block lying flat on the surface of the desk and lift it to a desired position.
\item Open sliding: Open a sliding door from left to right.
\item Close sliding: Close a sliding door from right to left.
\item Drawer: Open a closed desk drawer.
\item Close Drawer: Close an open desk drawer.
\item Sweep object: Sweep a block from the desk into an open drawer.
\item Knock object: Knock an upright object over.
\item Push red button: Push a red button inside a desk shelf.
\item Push green button: Push a green button inside a desk shelf.
\item Push blue button: Push a blue button inside a desk shelf.
\item Rotate left: Rotate a block lying flat on the table 90 degrees counter clockwise.
\item Rotate right: Rotate a block lying flat on the table 90 degrees clockwise.
\item Sweep left: Sweep a block lying flat on a table a specified distance to the left.
\item Sweep right: Sweep a block lying flat on a table a specified distance to the right.
\item Put into shelf: Place a block lying flat on a table into a shelf.
\item Pull out of shelf: Retrieve a block from a shelf and put on the table.
\end{itemize}

\subsubsection{Training Data}
An updated version of the Mujoco HAPTIX system is used to collect teleoperation demonstration data \cite{kumar2015mujoco}. Two types of demonstration data are collected for this experiment: 1) `play' data, collected without any specific task in mind but meant to cover many different possible object interactions, which is fed to \lmp and \gcbc and 2) segmented positive demonstrations of individual tasks (`open a door', `push a button', etc.), fed to the individual BC baseline models. Our environment
exposes arm and object position and orientations as observations to the agent. We model an 8-dof continuous action space representing agent end effector position, rotation, and gripper control.
See an example of the playground data collected in \afig{grid_playground} and an example of the positive behavioral cloning demonstrations in \fig{grid_sliding_demo}.
We collected around 3 hours total of playground data and 100 positive demonstrations each of 18 tasks (1800 demonstrations total). We collect 10 positive demonstrations of each task to use for validation and 10 for test. Tasks are specified to goal-conditioned models by resetting the environment to the initial state of the demonstration, and feeding in the final state as the goal to reach.

\subsection{Results Details}

\subsubsection{Robustness to Perturbations}
\label{sec:appendix.robustness}

In \fig{robustness} and \afig{robustness_per_task}, we see how robust each model is to variations in the environment at test time. To do so, prior to executing trained policies, we perturb the initial position of the robot end effector. We find that the performance of policies trained solely from positive demonstration degrades quickly as the norm of the perturbation increases, and in contrast, models trained on play data are more robust to the perturbation. We attribute this behavior to the well-studied ``distribution drift" problem in imitation learning (\citet{ross2011dagger}). Intuitively, models trained on expert demonstrations are susceptible to compounding errors when the agent encounters observations outside the expert training distribution. In interpreting these results we posit 1) the lack of diversity in the expert demonstrations allowed policies to overfit to a narrow initial starting distribution and 2) a diverse play dataset, with repeated, non-stereotyped object interaction and continuous collection, has greater coverage of the space of possible state transitions. This would make it more difficult for an initial error (or perturbation) to put the agent in an observation state outside its training distribution, ameliorating the compounding drift problem.

\begin{figure*}[h]
\begin{center}
\centerline{\includegraphics[width=\linewidth]{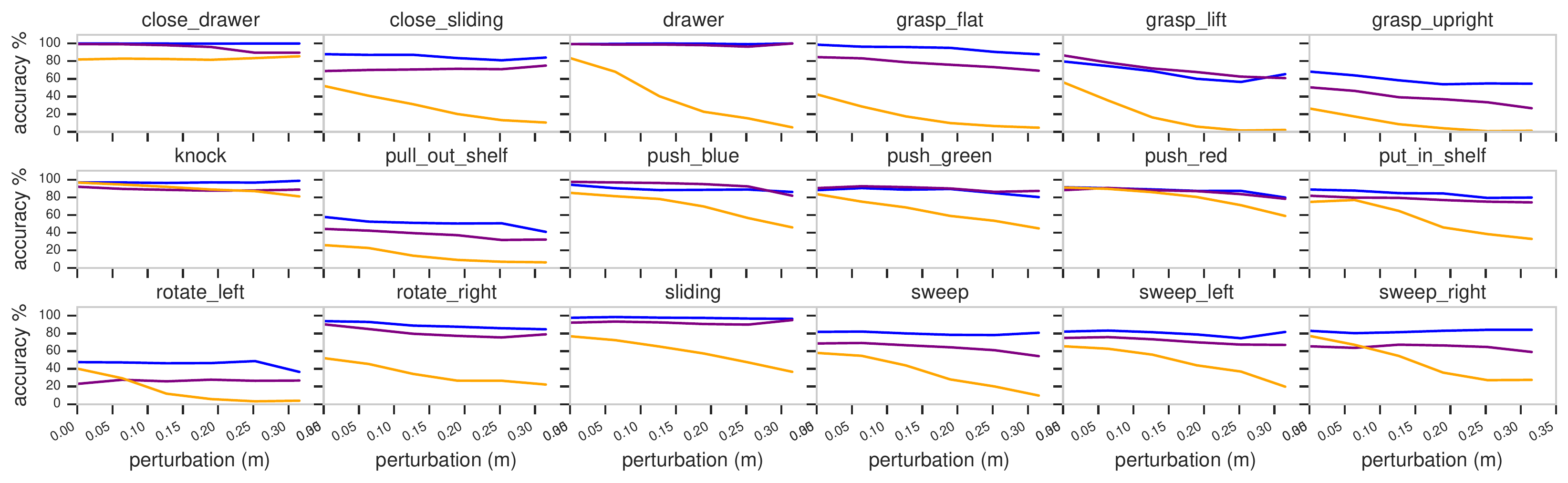}}
\caption{\textbf{Success per task while perturbing starting position.} See \fig{robustness} for the success averaged over all tasks. Perturbations vary (shown along the x-axis) between 0.0 to 0.4 meters from the initial position. We evaluate 3 models: Play-LMP (blue) and Play-GCBC (purple) are trained on play data, while BC (yellow) is trained on expert demonstrations. We find that models trained on play data are more robust to perturbations of the initial state.
}
\label{fig:appendix.robustness_per_task}
\end{center}
\end{figure*} 

\subsubsection{Emergent Retrying Behavior}
\label{sec:appendix.retry}

We find qualitative evidence that play-supervised models make multiple attempts to retry the task after initial failure. In \afig{retry_close_sliding} we see an example where our \lmp model makes 3 attempts to close a sliding door before finally achieving it. Similarly in \afig{retry_grasp_upright}, we see that the \lmp model, tasked with picking up an upright object, moves to successfully pick up the object it initially had knocked over. We find that this behavior does not emerge in models trained solely on expert demonstrations. We posit that the unique ``coverage" and ``incompletely functional" properties of play lend support to this behavior. A long, diverse play dataset covers many transitions between arbitrary points in state space. We hypothesize despite initial errors at test time lead the agent off track, it might still have (current state, goal state) support in a play dataset to allowing a replanning mechanism to succeed. Furthermore, the behavior is ``incompletely functional"---an operator might be picking a block up out of a drawer, accidentally drop it, then pick it right back up. This behavior naturally contains information on how to recover from, say, a ``pick and place" task. Furthermore, it would discarded from an expert demonstration dataset, but not a play dataset. 

\begin{figure*}[h]
\begin{center}
\centerline{\includegraphics[width=\linewidth]{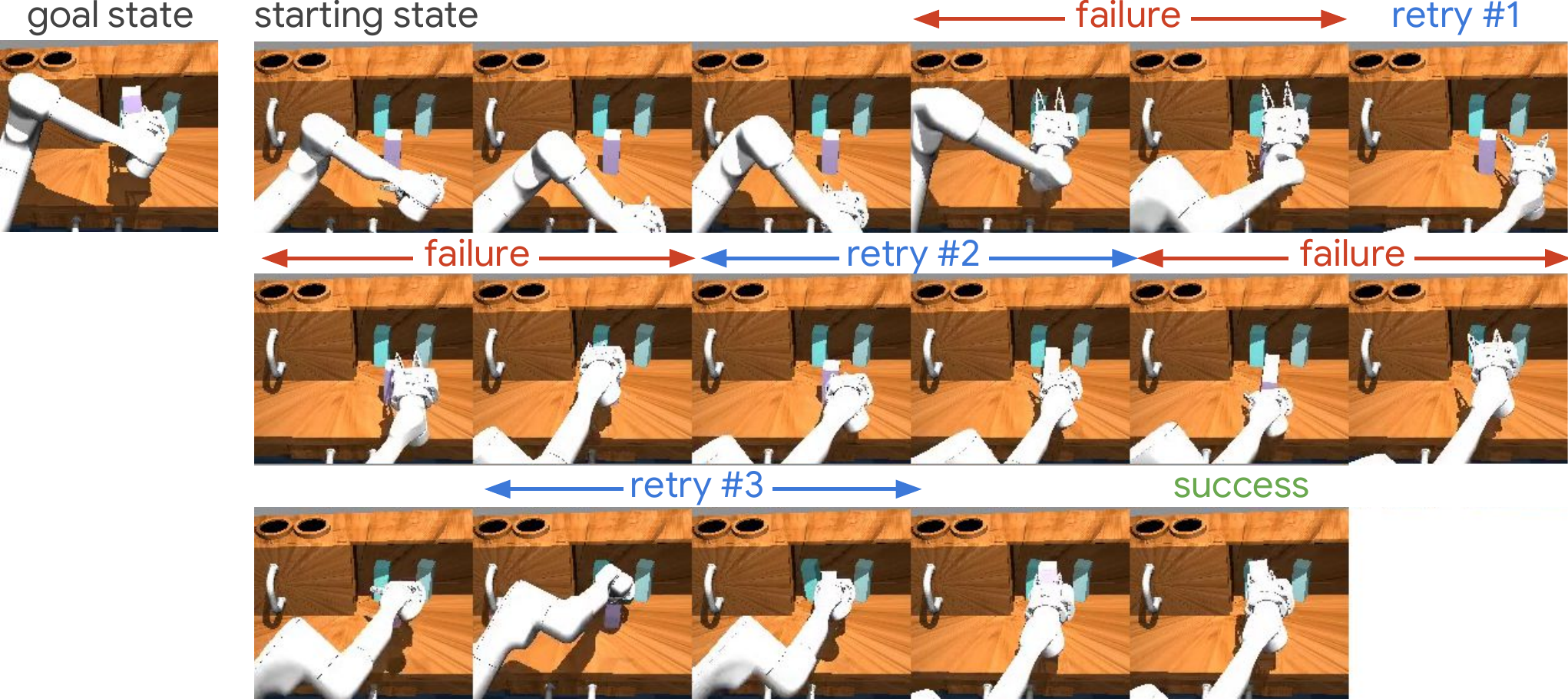}}
\caption{\textbf{Naturally emerging retrying behavior:} example run of Play-LMP policy on "grasp upright" task (grasping an object in upright position).
The agent fails initially, missing the block at first then knocking it over, then recovers successfully--picking up the knocked over block.
}
\label{fig:appendix.retry_grasp_upright}
\end{center}
\end{figure*} 

\begin{figure*}[h]
\begin{center}
\centerline{\includegraphics[width=\linewidth]{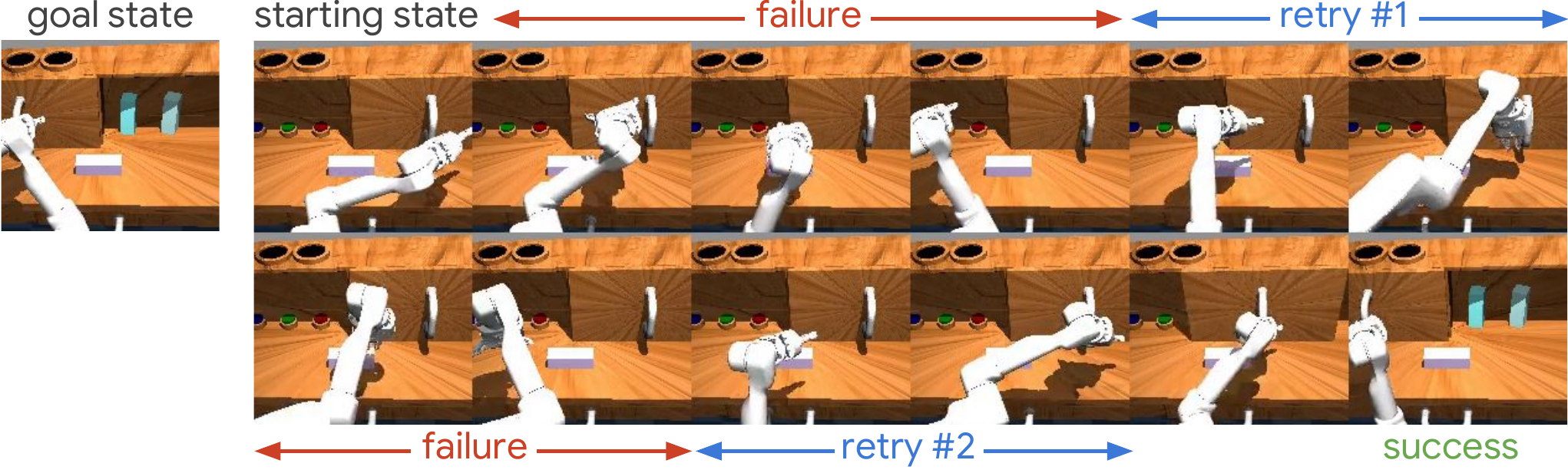}}
\caption{\textbf{Naturally emerging retrying behavior:} example run of \lmp policy on "close sliding" task (sliding door left to right).
The policy is aiming the reach the goal state (left), fails multiple times but retries without being explicitly asked to and is successful at the 3rd attempt.
}
\label{fig:appendix.retry_close_sliding}
\end{center}
\end{figure*} 

\subsubsection{Coverage Analysis of Interaction Space}
\label{sec:appendix.coverage}

In \fig{coverage_one} and \fig{coverage_seven}, we quantitatively measure the coverage of interaction space for different methods. To compute regions of interaction space, we quantized the 11 dimensions of action space corresponding to object interactions: the 3 position and 3 euler angle rotation coordinates of the block, the 1-d continuous {door open-close sensor, drawer open-close sensor, and 3 button sensors} into 10 bins each. During replay of the data, we counted the unique number of quantized bins visited by each of the three collection methods. 

The different collection methods plotted are:
“Expert demonstrations”, Fig 2c and 2d: This corresponds exactly to the BC baseline’s 18-task expert demonstration training data (90 minutes total, 100 demonstrations per task).
“Play data”, Fig 2c: This is the first 90 minutes of the 7h play dataset, restricted to the same size as the expert data for fair comparison. 
“Play data”, Fig 2d: This is our largest 7h play dataset, used to train our pixel experiment models. Our
state models were trained on a smaller dataset, up to 180 minutes of play (see Fig 8).
"Random": we collected a random exploration dataset in the environment by sampling actions uniformly from the bounds of the allowed action space.

Plots were generated by iterating the respective datasets and keeping track of summed time (x-axis) and cardinality of the set of visited quantized interaction space bins (y-axis).


\subsection{Limitations}
\label{sec:appendix.limitations}

Like other methods training goal-conditioned policies, we assume tasks important to a user can be described using a single goal state. This is overly limiting in cases where a user would like to specify \emph{how} they want the agent to do a task in addition to the the desired outcome, e.g. ``open the drawer slowly." As mentioned earlier, we could in principle use the trained sequence encoder \venc to perform this type of full sequence imitation. We hope to explore this in future work.

The scope of this work is to show that in a single environment, individual task-agnostic models trained using self-supervision on cheap play data can be competitive with many expert-trained models trained on expensive demonstrations. We emphasize that this sort of “single-room generalization” is consistent with the traditional assumptions of imitation learning---that training and test tasks are drawn independently from the same distribution. For play-supervised models, tasks are indexed by the (current, goal) pair. This means we expect to generalize to test-time tasks, also indexed by (current, goal), that are ``close" to those seen during training. The question of out-of-distribution generalization---say, playing in the living room and generalizing to the kitchen---is left to future work.

We additionally make the assumption that play data is not overly imbalanced with regards to one object interaction versus another. That is, we assume the operator does not simply choose to play with one object in the environment and never the others. This is likely a brittle assumption in the context of lifelong learning, where an agent might prefer certain play interactions over others. In future work, we look to relax this assumption.

Finally, we parameterize the outputs of both \venc and \cgenc as simple unimodal gaussian distributions for simplicity, potentially limiting the expressiveness of our latent plan space. Since Play-LMP can be interepreted as a conditional variational autoencoder, we might in future work consider experimenting with lessons learned from the variational autoencoder literature, for example more flexible variational posteriors, discrete rather than continuous codes in latent plan space (\citet{van2017neural}), etc.


\end{document}